\documentclass[11pt,a4paper]{article}
\usepackage{times,latexsym}
\usepackage{url}
\usepackage{listings}
\usepackage{graphicx}
\usepackage[T1]{fontenc}

\usepackage{multirow}
\usepackage[normalem]{ulem}
\useunder{\uline}{\ul}{}

\usepackage[table,xcdraw, dvipsnames]{xcolor}
\definecolor{verylightgray}{gray}{0.925}

\definecolor{plotly_blue}{HTML}{636EFA}
\definecolor{plotly_green}{HTML}{00CC96}
\definecolor{plotly_red}{HTML}{EF553B}

\usepackage{pifont}
\newcommand{\tick}{\textcolor{green!75!black}{\ding{51}}}%
\newcommand{\cross}{\textcolor{red}{\ding{55}}}%

\usepackage[acceptedWithA]{tacl2021v1}

\usepackage{xspace,mfirstuc,tabulary}

\usepackage{hyperref} 

\newif\iftaclinstructions
\taclinstructionsfalse 
\iftaclinstructions
\renewcommand{\confidential}{}
\renewcommand{\anonsubtext}{(No author info supplied here, for consistency with
TACL-submission anonymization requirements)}
\newcommand{\instr}
\fi

\iftaclpubformat 

\else

\fi

\title{Position: On the Methodological Pitfalls of Evaluating Base LLMs for Reasoning}

\author{
  Jason Chan 
  \qquad
  Zhixue Zhao
  \qquad
  Robert Gaizauskas
  \ \\
  University of Sheffield, UK
  \\
  \texttt{ \{jlychan1, r.gaizauskas, zhixue.zhao\}@sheffield.ac.uk}
}

\date{}

\begin{document}
\maketitle
\begin{abstract}

Existing work investigates the reasoning capabilities of large language models (LLMs) to uncover their limitations, human-like biases and underlying processes. Such studies include evaluations of base LLMs (pre-trained on unlabeled corpora only) for this purpose. Our position paper argues that evaluating base LLMs' reasoning capabilities raises inherent methodological concerns that are overlooked in such existing studies. We highlight the fundamental mismatch between base LLMs' pretraining objective and normative qualities, such as correctness, by which reasoning is assessed. In particular, we show how base LLMs generate logically valid or invalid conclusions as coincidental byproducts of conforming to purely linguistic patterns of statistical plausibility. This fundamental mismatch challenges the assumptions that (a) base LLMs' outputs can be assessed as their \textit{bona fide} attempts at correct answers or conclusions; and (b) conclusions about base LLMs' reasoning can generalize to post-trained LLMs optimized for successful instruction-following. We call for a critical re-examination of existing work that relies implicitly on these assumptions, and for future work to account for these methodological pitfalls.

\end{abstract}

\section{Introduction}

Reasoning is the process of drawing conclusions based on existing information (\citealp{oxford_handbook_intro_ch}; \citealp{mondorf2024beyond}). It is a cornerstone of intelligence that has traditionally been the subject of extensive study in philosophy and psychology. In artificial intelligence, systems as early as the Logic Theorist \cite{newell1956logic} and General Problem Solver \cite{newell_gps} have been proposed to simulate human reasoning processes using rules in formal logic.

This longstanding pursuit of artificial systems that can simulate reasoning continues to this day, as the reasoning capabilities and behavior of large language models (LLMs) have grown into a highly active area of current research (\citealp{mondorf2024beyond}; \citealp{ke2025a}, etc.). However, in contrast to earlier symbolic systems designed expressly to simulate reasoning, LLMs' ability to reason has been characterized as an ``\textit{emergent}'' property (\citealp{wei2022emergent}; \citealp{berti2025emergentabilitieslargelanguage})\footnote{But see \citet{schaeffer2023are} and \citet{lu-etal-2024-emergent} etc. for differing views.} that can be ``\textit{elicited}'' by prompting \cite{10.5555/3600270.3602070_cot_reasoning}.

Accordingly, current work commonly rests on a key assumption: \textit{the reasoning behavior and performance of an LLM can be evaluated by assessing its outputs in response to the appropriate prompts, regardless of whether the model itself has undergone any form of post-training to optimize it for instruction-following and correct reasoning}. On this basis, comparisons among different model sizes and types (e.g. pre-trained only versus instruction-tuned) are typically treated as only a matter of degree in capability, e.g. with instruction-tuning often described as a means of \textit{improving} a model's reasoning capabilities \cite{10.5555/3722577.3722647}. This premise underlies a wide range of studies that have experimented wholly or partly with base LLMs to draw conclusions about systemic reasoning errors, human-like biases, and specific ``circuits'' believed to explain a model's reasoning behavior (see Section \ref{sec:llm_reasoning_lit_review}). 

In this position paper, we challenge this foundational assumption and argue that \textbf{the practice of evaluating base LLMs for reasoning raises a number of inherent methodological issues overlooked in existing work.} Specifically:

\begin{enumerate}
    \item We highlight a fundamental mismatch in that base LLMs are optimized for linguistic plausibility instead of correctness or other similar desirable normative qualities that reasoning tasks are designed to evaluate (Section~\ref{sec:main_argument}); 

    \item We argue and empirically demonstrate how base LLMs conforming to simple linguistic patterns can generate logically valid or invalid conclusions as purely coincidental byproducts of optimizing for linguistic plausibility (Section~\ref{sec:case_study}); 

    \item We argue that the ambiguity of whether base LLMs' outputs are \textit{bona fide} attempts at correct answers is a critical confounding variable in assessing these models for reasoning abilities. When uncontrolled, this confound undermines conclusions about the models' inherent reasoning limitations or systematic biases (Section~\ref{sec:implications}); and

    \item Finally, we argue that findings on base LLMs' reasoning cannot be assumed to generalize to instruct LLMs, as they optimize for fundamentally different objectives. We therefore call for such claims in existing studies to be critically re-examined (Section~\ref{subsec:problem_generalization}).
\end{enumerate}

\section{Training LLMs}\label{sec:training_llms}

A typical LLM training pipeline (e.g. \citealp{grattafiori2024llama3herdmodels}; \citealp{abdin2024phi4technicalreport}; \citealp{lambert2025tulu3pushingfrontiers}) broadly consists of two stages: pre-training and post-training. 

In \textbf{pre-training}, a model is trained on large text corpora to maximize the conditional probability of the next token given its preceding context. These corpora include general web pages, textbooks, academic papers, code repositories and synthetically generated datasets, with web pages typically occupying a large portion of the data mix (e.g. \citealp{10.5555/3618408.3618510}; \citealp{olmo20252olmo2furious}).  

In certain cases, pre-training itself is split into two sub-stages, such that, in the latter sub-stage (commonly referred to as ``\textit{mid-training}''), the model is trained on a smaller subset of unlabeled data that is typically domain-specific (e.g. math, code) and of a higher quality (see e.g. \citealp{feng2024maximizedataspotentialenhancing}; \citealp{blakeney2024does}; \citealp{olmo20252olmo2furious}). 

We refer to models that have been pre-trained only, without any post-training, as \textit{base LLMs}.

Broadly speaking, the goal of \textbf{post-training} is to optimize a base LLM to follow user instructions, which includes correctly answering questions and successfully completing tasks specified in prompts to the model. For this purpose, post-training typically consists of multiple sub-stages, such as instruction-tuning (IT) \cite{instruction_tuning} and preference-tuning (PT) \cite{preference_tuning},  which differ among model families. 

In IT, the model is likewise trained to maximize the conditional probability of the next token given its preceding context, but with three key differences illustrated by the following sequence\footnote{Example from \citet{zhao2024wildchat}.}: 
\vspace{0.125in} 

\colorbox{verylightgray}{%
\begin{minipage}{0.43\textwidth}\small
\texttt{\textcolor{red}{<|endoftext|><|user|>\textbackslash n Explain University level Introductory Statistics to me like I'm a child\textbackslash n
<|assistant|>\textbackslash n}\textcolor{blue}{Okay little buddy, imagine you have a big bag of differently colored candies ... <|endoftext|>}}
    \end{minipage}
}

\begin{enumerate}
    \item Each instance in the IT data is a pair of text sequences: a \textcolor{red}{user instruction} and a \textcolor{blue}{target response} to that instruction.\footnote{For our purposes, we can disregard the use of system prompts and multi-turn sequences.} 
    \item Each pair of text sequences is formatted with a chat template (which demarcates instructions with special tokens e.g. ``\textit{<|user|>}'') and concatenated into a single sequence.
    \item Only the loss corresponding to tokens in the \textcolor{blue}{target response} sequence is taken into account, whereas loss corresponding to the \textcolor{red}{user instruction} sequence is masked.
\end{enumerate}

In PT, the model is further optimized using datasets with each instance consisting of an instruction prompt, a preferred response and a rejected response (see e.g. \citealp{bai2022traininghelpfulharmlessassistant}). While various techniques such as direct preference optimization \cite{dpo_10.5555/3666122.3668460} are introduced for this purpose, a common objective is for the model to maximize the difference between the log likelihood assigned to the preferred response and that assigned to the rejected response. This step ensures that outputs generated by the model are desirable according to human judgments\footnote{Or, in LLM-as-a-judge \cite{10.5555/3666122.3668142} or RLAIF \cite{lee2024rlaifvsrlhfscaling} approaches, judgments of existing LLMs that have themselves already been preference-tuned.} based on a specified set of normative criteria, such as ``\textit{helpfulness}”, ``\textit{instruction following}”, ``\textit{honesty}” and ``\textit{truthfulness}” \cite{lambert2025tulu3pushingfrontiers}.

We refer to models that have undergone a post-training stage\footnote{We interpret \textit{post-training} broadly to include any method of training whose goal is to optimize a model to follow user instructions in some desirable manner e.g. successfully, helpfully etc., which includes techniques such as applying reinforcement learning (with correct answers as reward signals) directly to base LLMs \cite{deepseekai2025deepseekr1incentivizingreasoningcapability} without going through any fine-tuning stages such as IT as we described.} as \textit{instruct LLMs}.

\section{Current Work in LLM Reasoning}\label{sec:llm_reasoning_lit_review}

\begin{table*}[]
\footnotesize
\centering
\begin{tabular}{llllcc}
\hline
\textbf{Category} & \textbf{Subcategory} & \textbf{Study} & \textbf{\begin{tabular}[c]{@{}l@{}}Reasoning \\ domain/type\end{tabular}} & \multicolumn{1}{l}{\textbf{\begin{tabular}[c]{@{}l@{}}Base \\ LLMs?\end{tabular}}} & \multicolumn{1}{l}{\textbf{\begin{tabular}[c]{@{}l@{}}Instruct \\ LLMs?\end{tabular}}} \\ \hline
\multirow{14}{*}{\begin{tabular}[c]{@{}l@{}}Evaluating \\ capabilities\\ and limitations\end{tabular}} & \multirow{2}{*}{\begin{tabular}[c]{@{}l@{}}Comparative analysis\\ (base vs instruct)\end{tabular}} & \citealp{li-etal-2025-small-models} & General & \tick & \tick \\
 &  & \citealp{bigoulaeva2025inherentlimitspretrainedllms} & General & \tick & \tick \\ \cline{2-6} 
 & \multirow{2}{*}{\begin{tabular}[c]{@{}l@{}}Comparative analysis\\ (model scale)\end{tabular}} & \citealp{chen2024biggerdeeperbetterprobing} & General; math & \tick & \cross \\
 &  & \citealp{mckenzie2023inverse} & General & \tick & \tick \\ \cline{2-6} 
 & \multirow{3}{*}{Benchmark evaluation} & \citealp{srivastava2024functionalbenchmarksrobustevaluation} & Math & \tick & \tick \\
 &  & \citealp{jin2023cladder} & Causal & \tick & \tick \\
 &  & \citealp{yang-etal-2024-language} & Inductive & \tick & \cross \\ \cline{2-6} 
 & \begin{tabular}[c]{@{}l@{}}Benchmark evaluation;\\ error analysis\end{tabular} & \citealp{Griot2025} & Medical & \tick & \tick \\ \cline{2-6} 
 & \multirow{3}{*}{Error analysis} & \cite{ye2022the} & General & \tick & \tick \\
 &  & \citealp{zheng2024large} & General & \tick & \tick \\
 &  & \citealp{razeghi-etal-2022-impact} & Math & \tick & \cross \\ \cline{2-6} 
 & \multirow{3}{*}{Enhancement method} & \citealp{huang-etal-2023-large} & General & \tick & \cross \\
 &  & \citealp{sharma2024the} & General & \tick & \cross \\
 &  & \citealp{hao2025training} & General & \tick & \cross \\ \hline
\multirow{4}{*}{\begin{tabular}[c]{@{}l@{}}Comparisons with\\ human reasoning\end{tabular}} & \multirow{4}{*}{\begin{tabular}[c]{@{}l@{}}Cognitive bias and\\ error analysis\end{tabular}} & \citealp{10.1093/pnasnexus/pgae233} & General & \tick & \tick \\
 &  & \citealp{Yax2024_base_rate_fallacy} & General & \tick & \tick \\
 &  & \citealp{doi:10.1073/pnas.2218523120} & General & \tick & \tick \\
 &  & \citealp{eisape-etal-2024-systematic} & Logic & \tick & \cross \\ \hline
\multirow{5}{*}{\begin{tabular}[c]{@{}l@{}}Mechanistic\\ interpretability\end{tabular}} & \multirow{5}{*}{\begin{tabular}[c]{@{}l@{}}Reasoning circuit /\\ pathway analysis\end{tabular}} & \citealp{hou-etal-2023-towards} & General; CoT & \tick & \cross \\
 &  & \citealp{dutta2024how} & Logic; CoT & \tick & \cross \\
 &  & \citealp{kim-etal-2025-reasoning} & Logic & \tick & \cross \\
 &  & \citealp{stolfo-etal-2023-mechanistic} & Math & \tick & \cross \\
 &  & \citealp{lee2024causal} & Causal & \tick & \cross \\ \hline
\end{tabular}
\caption{Examples of existing representative work in LLM reasoning. As shown in the second rightmost column, all the listed studies conduct experiments with base LLMs on the \textbf{assumption that these models can be prompted to engage in reasoning}. As shown in the rightmost column, relying on this assumption, some studies compare the reasoning of base LLMs with instruct LLMs, while others conduct experiments exclusively using base LLMs.}
\label{tab:llm_reasoning_studies}
\end{table*}

The practice of evaluating base LLMs on reasoning is prevalent in existing work. As we show in this Section with representative studies, this is typically carried out with two implicit assumptions: 
\begin{enumerate}
    \item When prompted, base LLMs are genuinely attempting to answer correctly as per the task instructions; and 
    \item Findings from base LLMs (e.g. on their reasoning limitations) can be generalized to support claims about LLMs as a whole, including instruct LLMs.
\end{enumerate}
While we will challenge both these assumptions in subsequent Sections, we first establish the context for our critique by illustrating its necessity in light of current research practices.

\textbf{Capability evaluation}. Numerous studies have evaluated LLMs' ability to reason with logic \cite{liu2025logicalreasoninglargelanguage}, mathematics \cite{ahn-etal-2024-large}, commonsense (\citealp{li-etal-2022-systematic}; \citealp{chan2025rulebreakers}), specialist and scientific knowledge \cite{zhang-etal-2024-comprehensive-survey} across various domains (see e.g. \citealp{yu_llm_survey} and \citealp{giadikiaroglou-etal-2024-puzzle} for further surveys). As shown in Table \ref{tab:llm_reasoning_studies}, studies typically assume that reasoning can be elicited from base LLMs as well as instruct LLMs, and that their capabilities differ only in degree (i.e. which type of LLM reasons \textit{better}) and not in nature (see e.g. \citealp{li-etal-2025-small-models}). Based on this assumption, studies have also conducted experiments exclusively on base LLMs to draw conclusions about reasoning limitations of LLMs \textit{in general} (see Table \ref{tab:llm_reasoning_studies}). 

\textbf{Comparisons with human reasoning}. Similarly, the assumption that base LLMs are reasoning and genuinely attempting to generate correct answers or conclusions is also relied upon by studies that compare the reasoning behavior of LLMs against that of human reasoners. Existing work in this category has found that LLMs exhibit certain human-like cognitive biases, such as content effect \cite{10.1093/pnasnexus/pgae233}, base rate fallacy \cite{Yax2024_base_rate_fallacy} and figural effect \cite{eisape-etal-2024-systematic}. 

\textbf{Behavioral understanding}. A growing body of interpretability research aims to understand the internal processes behind LLM reasoning. One commonly adopted approach is mechanistic interpretability \cite{olah2020zoom}, which aims to identify specific computational subgraphs (or ``circuits'') that causally explain how a model computes its outputs when prompted with certain contexts or premises (e.g. \citealp{wang2023interpretability}; \citealp{mech_interp_logic}; \citealp{nikankin2025arithmetic}). As shown in Table \ref{tab:llm_reasoning_studies}, such mechanistic analyses have typically been conducted only on base LLMs.

Circuits are typically identified using \textit{activation patching} (\citealp{vig_activation_patching}; \citealp{zhang2024towards}), which involves pairs of minimally differing prompts, \textit{clean} versus \textit{corrupt}. For example, to identify the circuit responsible for a model's capability for syllogistic reasoning, \citet{kim-etal-2025-reasoning} uses clean prompts in the form of ``\textit{All A are B. All B are C. Therefore, all A are}''\footnote{In this case, a model capable of syllogistic reasoning is expected to predict ``C'' as the most likely next token.} and corrupt prompts ``\textit{All A are B. All \underline{\textbf{Q}} are C. Therefore, all A are}''. First, the model is run on both the clean and corrupt prompts, with its respective output logits recorded. In the clean run, intermediate output values of selected model components (e.g. individual attention heads at different model layers) are also stored. Then, when the model is run again on the corrupt prompt, the value of a particular component is overwritten by ``patching in'' the corresponding stored value from the clean run, and the resulting change in the model's output logits is recorded. By systematically patching each model component, researchers can identify those that result in large changes, which would indicate that these components play a key role in the circuit for the particular model behavior being studied.

\section{The Fundamental Mismatch in Evaluating Base LLMs on Reasoning}\label{sec:main_argument}

Our position recognizes but sidesteps the contentious theoretical debate on whether LLMs are capable of \textit{genuine} reasoning, and the philosophical issues with characterizing what this entails (see e.g. \citealp{10.1145/3624724}; \citealp{cappelen2025goinghogphilosophicaldefense}). Instead, we focus on the practical issue of \textit{evaluation methodology}. Specifically, we argue that the practice of experimenting on base LLMs by prompting in order to draw conclusions about the reasoning limitations and processes of LLMs \textit{in general} (whether base or instruct) is problematic. 

\subsection{Core Principle: Performance Measures should Reflect Models' Objectives}\label{subsec:analogy}
A core principle in machine learning is that the performance measures on which a model is evaluated should reflect the tasks that the model is learning to solve (\citealp{10.5555/541177_ml_book}; \citealp{10.5555/3086952_dl_book_goodfellow}). For example, consider a set of medical image classifiers designed and trained only for classifying benign versus malignant tumors. If we evaluate these models on how accurately they predict the gender of the patients, we are evaluating them on a task they were never optimized for. Suppose we find these models perform far worse at predicting the gender of older patients than younger ones. Can we conclude that this type of classifier generally exhibits bias in gender prediction?

We argue that this conclusion is unwarranted. A prediction bias is a systematic error \textbf{as measured against the model's targets} and not against an arbitrary variable that the model was not trained to predict in the first place. In this case, there is a fundamental mismatch between the task being evaluated (gender prediction) and the model's objective (tumor classification). In other words, any capability the model has to predict gender is \textit{purely incidental to its objective}. As such, the finding cannot support the conclusion that, even if models of this type were \textit{specifically} trained or fine-tuned for gender prediction, they would exhibit the same systematic bias in performing worse at predicting the gender of older patients than younger ones.

\subsection{Base LLMs Optimize for Linguistic Plausibility, not for Normative Qualities}

As described in Section \ref{sec:training_llms}, a base LLM's objective optimizes only for linguistic plausibility and not for normative qualities, such as correctness or logical validity, by which we typically assess reasoning. In other words, base LLMs are not trained for the purpose of producing outputs that conform to our standards for what counts as ``good'' reasoning; instead, they are trained only for the purpose of producing outputs that are statistically the most likely continuation of a given context.\footnote{We use the term ``\textit{linguistic plausibility}'' in this specific purely statistical sense (as opposed to e.g. \textit{grammaticality}).} 

Applying the analogy in Section \ref{subsec:analogy}, the heart of our argument is that \textbf{the ability of base LLMs to produce outputs that reflect good reasoning is therefore purely incidental to its objective in optimizing for linguistic plausibility}.

\subsection{Why High-Quality Pretraining Data does not Address the Mismatch}\label{subsec:high_quality_data}
Here, a key objection to this position is that by curating high-quality pretraining corpora, base LLMs are already \textit{implicitly} optimizing for correctness and other desirable normative qualities associated with reasoning. We argue that this view misinterprets both the nature of high-quality data and the objective of the base LLMs. 

While corpora typically considered as high-quality (such as human-written academic textbooks) can consist of example questions (or premises) followed by correct answers (or valid conclusions), they can also contain \textit{bona fide} demonstrations of incorrect answers and flawed reasoning. This is by no means an issue of poor data quality, misinformation or bias, the effects of which have all been discussed in existing work (e.g. \citealp{lin-etal-2022-truthfulqa}; \citealp{mckenzie2023inverse}). On the contrary, these demonstrations of incorrect answers and reasoning are a valuable feature in high-quality educational texts that serve a clear pedagogical purpose.\footnote{In the pretraining corpus olmo-mix-1124 \cite{olmo20252olmo2furious}, for example, we identified extracts of \textbf{incorrect} reasoning processes intended as demonstrations of e.g. a ``\textit{common mistake in applying the demand and supply framework}'' (proofpile-shard-0035-31144), and how an incorrect solution to a ``\textit{Bayesian inference problem}'' might be mistakenly derived (proofpile-shard-0035-24997).} 

For instance, a textbook might include demonstrations in the format of ``\textit{Question: [...] Answer: [...]}'', followed by a helpful explanation of why the answer is incorrect. Reproducing this purely statistical pattern, a base LLM prompted with a multiple-choice question might continue the context ``\textit{Question: [...] Answer: }'' not with what it ``believes'' to be correct answer, but with an \textit{incorrect} answer accompanied by an explanation, e.g. ``\textit{C. The answer is incorrect because...}''. 

What this shows is that pre-training a model on high-quality data does not on its own warrant the claim that the model's outputs are therefore optimizing for correctness or validity as opposed to mere linguistic plausibility.

\subsection{Correctness and Normative Qualities are Valid Measures for Instruct LLMs}

As recognized by \citet{preference_tuning}, the task of language modeling (predicting the most likely continuation of a preceding context, given an unlabeled pretraining corpus) is distinct from the task of helpfully and safely following user instructions. This distinction is what motivates instruction-tuning, so that models like InstructGPT \cite{preference_tuning} are re-optimized for the latter task. We argue that this re-optimization is precisely what entitles us to evaluate instruct LLMs for normative qualities such as correctness in their generated outputs.

For example, as explained in Section \ref{sec:training_llms}, instruct LLMs are fine-tuned on instruction-response pairs where the response is expected to \textit{always} demonstrate a \textit{successful fulfillment} of the instruction. This data is formatted with a chat template using unique markers such as ``\textit{<|user|>}'' and ``\textit{<|assistant|>}'' that are not expected to have appeared anywhere in the model's pretraining data.

Because the text following ``\textit{<|assistant|>}'' is always a demonstration of \textit{successful} instruction-following, we are deliberately signaling to the model that its own generated continuation must also be a successful fulfillment. This includes correctly answering a question or drawing a logical conclusion, as opposed to simply demonstrating an unsuccessful response.\footnote{As discussed in Section \ref{sec:training_llms}, we recognize that such fine-tuning is only one of the ways to re-optimize the model for correctness and provide it with a clear signal to generate a correct answer.}

\textbf{This functional shift from language modeling to instruction-following means that generating correct and logical answers is no longer an incidental byproduct but a valid and important performance measure for how well a model optimizes for its new objective.} Just as humans apply reasoning to achieve goals in real-life scenarios \cite{EVANS1993165}, reasoning plays a clear instrumental role in enabling an instruct LLM to successfully fulfill a wide variety of user instructions, such as correctly answering complex questions or debugging code \cite{llms_debugging}. While the model is still mechanically performing next-token prediction, its re-optimization is intended to ensure that what the model considers to be linguistically plausible now \textbf{encapsulates} the normative qualities, such as correctness and helpfulness, that we expect from responses that successfully fulfill user instructions.

\section{Case Study: Debunking the Emergence of Logical Reasoning in Base LLMs}\label{sec:case_study}

We present a case study to explain how logically valid or invalid conclusions can emerge as coincidental byproducts from a base LLM optimized only for linguistic plausibility. We show how observations about these byproducts can lead to unwarranted claims about the model's underlying reasoning capabilities and limitations, demonstrating the methodological issues we set out in the previous Section.

\subsection{Mechanisms of Linguistic Plausibility}

Existing work has identified a number of behavioral patterns in base LLMs and the circuits that mechanically explain such behaviors (\citealp{olsson2022context}; \citealp{wang2023interpretability}; \citealp{hanna2023how} etc.). We isolate two circuits in particular:

The \textbf{Indirect Object Identification} (IOI) circuit \cite{wang2023interpretability} explains how a model predicts that a context in the form of ``\textit{When John and Mary went to the store, John gave a bottle to}” should be followed by ``\textit{Mary}”. The circuit implementing this behavior:
\begin{enumerate}
    \item identifies all the names that have appeared in the context (``\textit{Mary}” and ``\textit{John}”);
    \item suppresses the name which has already been repeated in the context (``\textit{John}”); and
    \item outputs the remaining name (``\textit{Mary}”).
\end{enumerate}

An \textbf{induction head} (IH) circuit \cite{olsson2022context} explains how a model predicts that a context of the form ``\textit{[A][B]....[A]}” should be followed by ``\textit{[B]}”. For example, if two tokens ``\textit{Harry Potter}” have appeared in the context, the model predicts that the next time the token ``\textit{Harry}” appears, it will likely be followed again by the token ``\textit{Potter}”. 

\begin{figure}
    \centering
    \includegraphics[width=0.425\linewidth]{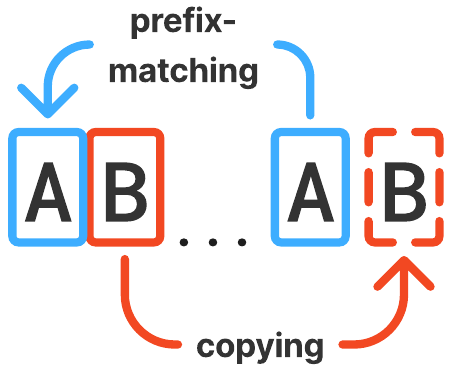}
    \caption{The two stages of an IH circuit explaining how a model predicts that the token ``\textit{[B]}'' follows the context ``\textit{[A][B]...[A]}''. ``...'' denotes other tokens.}
    \label{fig:ih_circuit}
\end{figure}
As Figure \ref{fig:ih_circuit} shows, the circuit for this behavior:
\begin{enumerate}
    \item identifies the previous occurrence of a token ``\textit{[A]}” in the context (\textit{prefix-matching}); and
    \item boosts the logit of the token that followed ``\textit{[A]}” when ``\textit{[A]}” previously occurred (i.e. ``\textit{[B]}'') (\textit{copying}).
\end{enumerate}

Note that, as described by these circuits, \textbf{the model is predicting whether a particular token follows a given context as a matter of linguistic plausibility based solely on the syntactic properties of that token} (e.g. whether and where it has previously appeared in the context, whether it is a name). Its mechanisms are indifferent to whether the generated tokens constitute a ``correct” or ``logically valid” continuation of the context. 

\subsection{Logically Valid Conclusions as Incidental Byproducts}\label{subsec:valid_conclusions_are_byproducts}

Despite being purely linguistic, we now show how these circuits are \textbf{sufficient} to generate tokens constituting logically valid conclusions in some cases.
 
Suppose a base LLM consists only of two circuits: IOI and IH. Given the four contexts below:
\begin{quote}
    E1: If Ann had lunch, then Sam made dinner. Ann had lunch. Therefore,

E2: Either Ann had lunch, or Sam made dinner. Ann did not have lunch. Therefore,

E3: Ann had lunch and Sam made dinner. Ann had lunch. Therefore,

E4: Ann had lunch if and only if Sam made dinner. Ann had lunch. Therefore,
    
\end{quote}

In each of E1 to E4, the model would generate ``\textit{Sam made dinner}.” which, in propositional logic \cite{language_proof_and_logic_propositional}, is a valid conclusion to the premises in the context. The mechanism is purely linguistic: 

\begin{enumerate}
    \item First, by the IOI circuit, the model can identify the two names that have appeared in the context, and output the only name that has not yet been repeated i.e. ``\textit{Sam}”.
    \item Then, by the IH circuit, the model generates ``\textit{made dinner.}” to follow the token ``\textit{Sam}”, since this sequence of tokens (``\textit{made dinner.}”) has followed ``\textit{Sam}'' when ``\textit{Sam}'' previously appeared in the context.
\end{enumerate} 

Based on the above, are we then justified in concluding that such an LLM has learned to perform logical reasoning in these scenarios, and that these circuits describe how they reason? We argue that no. The heart of our objection is that \textbf{the generation of correct or logically valid conclusions in these scenarios is merely a coincidental byproduct of a model optimized only for linguistic plausibility}. 

\subsection{Logically Invalid Conclusions as Incidental Byproducts}

The model's indifference to logical validity and correctness becomes clear when we present it with two further contexts below:

\begin{quote}
     E5: If Ann had lunch, then Sam made dinner. Sam made dinner. Therefore,
     
     E6: Either Ann had lunch, or Sam made dinner. Sam made dinner. Therefore,

\end{quote}

In both E5 and E6, ``\textit{Ann had lunch}” is a logically invalid conclusion. Yet, this is precisely what the model would generate \textbf{using exactly the same circuits} that had generated the logically valid conclusions in E1 to E4.

This illustrates that a model optimized solely for linguistic plausibility is indifferent to the validity or correctness of its generated tokens which we interpret to be a ``conclusion”. 

But can we claim that scenarios like E5 and E6 show merely that the model is prone to systematically committing certain kinds of logical fallacies when reasoning? We argue again that no: treating these outputs as genuine fallacies or ``reasoning errors” \textit{presupposes} that the model had aimed to generate correct or logically valid conclusions in the first place, which as we have argued cannot be established with respect to a model optimized simply for linguistic plausibility. 

\subsection{Empirical Demonstration}

In light of the above, we further conduct an experiment with existing base LLMs to empirically demonstrate this phenomenon. In doing so, we show that base LLMs responding indifferently with logically valid and invalid conclusions is not merely a theoretical conjecture but a real and quantifiable behavioral pattern.\footnote{We plan to publicly release the code and data used in this experiment.}

\textbf{Data}. We first define two templates, based on examples E1 and E5 in the previous subsections:

\begin{itemize}
    \item \textbf{Valid-form} (modus ponens): ``\textit{If Ann had lunch, then Sam made dinner. \underline{Ann had lunch}. Therefore,}''
    \item \textbf{Invalid-form} (affirming the consequent): ``\textit{If Ann had lunch, then Sam made dinner. \underline{Sam made dinner}. Therefore,}''
\end{itemize}

We generate 7,600 valid-form prompts by permuting the template with 20 pairs of first names (e.g. ``\textit{Ann}'', ``\textit{Sam}'')\footnote{Randomly sampled without replacement from a list of common names: \url{https://github.com/sigpwned/popular-names-by-country-dataset/}} and 380 pairs of predicates (e.g. ``\textit{had lunch}'', ``\textit{made dinner}''), which are unique combinations from a manually written list of 20 predicates. Then, for each valid-form prompt, we generate a corresponding \textit{minimally differing} invalid-form prompt by replacing the second premise with the other name and predicate in the context (as shown by the \underline{underlined segments} in the example prompt pair above), resulting in a total of 7,600 invalid-form prompts.

\textbf{Metric}. For each base LLM, we compute the proportion of all \textit{valid-form} prompts that the model completes with a target string; and the proportion of all \textit{invalid-form} prompts it completes with a target string. Importantly, the target string corresponding to each prompt (whether valid- or invalid-form) follows the identical linguistic pattern set out in Section \ref{subsec:valid_conclusions_are_byproducts}: starting with the name that is not duplicated in the context, followed by the tokens that had followed that name where it previously appeared in the context. For example, the target strings of the example valid- and invalid form prompts would be ``\textit{Sam made dinner.}'' and ``\textit{Ann had lunch.}'', respectively.\footnote{We deliberately include full stops as part of the target strings and require that an output \textbf{begin} with the target string to count as a match, so as to exclude conclusions of other forms e.g. ``\textit{Ann had lunch or he did not have lunch.}'' or ``\textit{we cannot say for certain whether Ann had lunch.}''}

\textbf{Models}. We test 13 open-weight base LLMs of different families, including Qwen2.5, OLMo 2 \cite{olmo20252olmo2furious}, Llama 3.1 and 3.2 \cite{grattafiori2024llama3herdmodels}, Qwen-2.5 \cite{qwen2025qwen25technicalreport} and Gemma 3 \cite{gemmateam2025gemma3technicalreport}, with model sizes ranging from 0.5B to 32B.

\textbf{Setup and implementation}. For all LLMs, we use the model version hosted on Hugging Face \cite{wolf-etal-2020-transformers} and run inference using vllm \cite{kwon2023efficient}. We prompt each model using greedy decoding, with temperature set to 0. All experiments are performed on NVIDIA A100 GPUs.

\begin{figure*}
    \centering
    \includegraphics[width=0.925\textwidth]{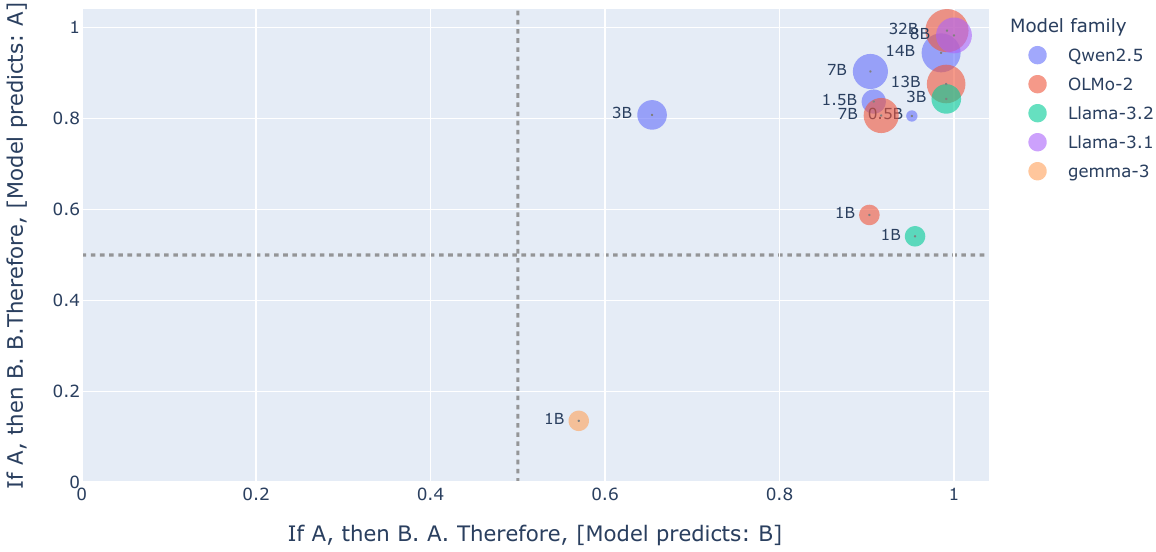}
    \caption{For each base LLM, we measure the proportion of \textit{valid-form} prompts the model completes with the target string (x-axis), against the proportion of \textit{invalid-form} prompts it completes with the target string (y-axis). Models appearing near the top-right corner reflect a strong tendency to generate the target strings in response to both valid- and invalid-form prompts. Dot size denotes model size.}
    \label{fig:experiment_plot}
\end{figure*}

\textbf{Results}. As shown in Figure \ref{fig:experiment_plot}, most base LLMs cluster towards the top-right corner of the top-right quadrant. This indicates that they tend to generate outputs that match the target strings in response to most prompts, regardless of whether they are valid-form (in which case the target string is a valid conclusion) or invalid-form prompts (in which case the target string is an invalid conclusion). \textbf{This behavior conforms to the hypothesized mechanism, and illustrates how ``valid'' and ``invalid'' conclusions generated by these models are equally incidental artifacts of a purely linguistic mechanism that is not optimizing for correctness or validity.}

While certain models are outliers to this general pattern (notably gemma-3-1b-pt in the lower-right quadrant), this by no means implies that they have ``correctly identified and avoided fallacies''. On the contrary, our manual inspection reveals that these models tend to produce other irrelevant or invalid conclusions that are highly similar to the target string, with a number of predictable patterns to these mismatches. For example, they tend to confuse the tense of the two predicates when they differ (e.g. generating ``\textit{Sam makes dinner}'' as opposed to ``\textit{Sam made dinner}''), and mix up tokens belonging to different predicates (e.g. generating ``\textit{Ann had dinner}'' when the predicates are ``\textit{had lunch}'' and ``\textit{made dinner}''). \textbf{As such, these outliers and systematic mismatches in fact support our overarching position that the models' outputs conform (albeit imperfectly) to patterns that are purely linguistic, without regard to validity or correctness.}

\section{Implications}\label{sec:implications}

The argument we have advanced thus far carries two key implications. First, evaluating base LLMs for reasoning comes with a number of inherent methodological problems, which we discuss in Sections \ref{subsec:unwarranted_assume_correctness} and \ref{subsec:confounding}. Second, findings from these evaluations cannot be assumed to generalize to instruct LLMs, which we explain in Section \ref{subsec:problem_generalization}.

\subsection{Unwarranted Assumption: Base LLMs are Aimed at Correct Outputs}\label{subsec:unwarranted_assume_correctness}

First, given the example case in Section \ref{subsec:high_quality_data} where a base LLM continues the context with an incorrect answer followed by a pedagogical explanation for why the answer is incorrect,\footnote{To be clear, this is only one of the potential ways that a base LLM might continue a context other than attempting a correct answer of the user's question. From anecdotal observations, it is entirely possible for base LLMs to also recast its answer as merely part of the context for another question e.g. ``\textit{C. Is the above Answer True or False? A. True. B. False.}''. Importantly, all these continuations are perfectly coherent and compatible even with context that includes clear instructions e.g. ``\textit{answer the question correctly}''.} it is clearly problematic to simply extract the initial token ``\textit{C}'', interpret this as what the model considers to be the correct answer, and make claims about model's reasoning limitations on this basis. 

Yet, this approach reflects a common practice in the automatic evaluation of LLM reasoning: programmatically extracting specific tokens (or token spans) and probabilities from the model's outputs, scoring these as the model's answers against the ground truth labels and then either stopping the model from generating any more tokens or disregarding the rest of the model's outputs.\footnote{See e.g. as implemented by \url{https://github.com/EleutherAI/lm-evaluation-harness}} In essence, this practice is problematic because it relies on an unwarranted assumption that a base LLM is always aiming to generate a correct answer or conclusion by continuing the context, when the model is optimized only for linguistic plausibility and not for correctness. 

\subsection{Recognizing the Instruction-Following Confound when Assessing Reasoning}\label{subsec:confounding}

Put plainly, we cannot draw valid conclusions about a base LLM's reasoning capabilities (e.g., what it can or cannot do, or what systematic errors or biases it makes) if we are not even sure that the model is trying to follow the instruction to produce a correct answer in the first place. From a statistical perspective, this ambiguity, given a base LLM's objective, introduces a critical confounding variable (which we refer to as the ``\textit{instruction-following confound}'') when we evaluate a model's reasoning capabilities.

In this sense, although by no means a guarantee, optimizing a model to successfully follow instructions (e.g. by instruction-tuning) is an important control for this confounding variable, and not just another method for ``improving performance''. Consequently, when an instruct LLM fails to reason correctly, we have a much stronger \textit{justification} for attributing the error to a genuine lack of capability or a systematic bias or error, as opposed to the possibility that the model simply did not follow the user's instructions or demonstrations.

\subsection{Findings about Base LLMs cannot be Assumed to Generalize to Instruct LLMs}\label{subsec:problem_generalization}

As our analogy in Section \ref{subsec:analogy} shows, findings about an ability that is merely an incidental byproduct in a model (one not optimized for that objective) cannot be validly generalized to a model that \textit{has} been explicitly optimized for that same objective. In case of LLMs, it is unwarranted to assume that any apparent reasoning biases or limitations observed by evaluating base LLMs (optimized for linguistic plausibility) will persist even after these models are re-optimized for correctness and other normative qualities we expect in models that successfully fulfill user instructions. Likewise, given this objective mismatch, it is also unwarranted to assume that the mechanisms by which a base LLM produces its outputs are the same ones it will use when it is re-optimized as an instruct LLM to generate outputs that are correct as opposed to merely linguistically plausible.

\section{Alternative Views}

\textbf{AV: The boundary between base and instruct LLMs is fuzzy from a data perspective.}

\textit{It is impossible to draw a clear and categorical distinction that base LLMs only optimize for linguistic plausibility whereas instruct LLMs optimize for correctness. Instruction-following datasets are often mixed, whether deliberately or otherwise, into the pretraining data for base LLMs (e.g. \citealp{bai2023qwentechnicalreport}). For example, mixing in datasets with questions followed by reasoning chains and correct answers was found to give base LLMs improved reasoning performance that persists after post-training \cite{akter2025frontloadingreasoningsynergypretraining}}.

We recognize this lack of a hard boundary from a data perspective. However, our argument is not that pretraining data contains no examples of correct reasoning or instruction-following, but rather that \textbf{it does not deliberately exclude examples of faulty reasoning, incorrect answers, or logical fallacies}. This makes web-scale pretraining data a noisy signal if the aim of the pretraining stage is genuinely to optimize the model for correctness and other normatively desirable qualities. By contrast, in post-training, there is a deliberate curation to select only for examples of correct instruction-following, reasoning and question-answering. This signal is further reinforced using through chat templates with unique markers, which explicitly frame the user-provided context as an instruction that expects successful fulfillment (as discussed in Section \ref{sec:training_llms}).

Given the above, compared to base LLMs, we have a much stronger warrant for arguing that instruct LLMs are genuinely being optimized for correctness as opposed to mere linguistic plausibility. The distinction is methodologically critical: this deliberate curation and signaling is precisely what justifies the claim that the instruction-following confound is controlled (at least to a sufficient degree) in instruct LLMs, so that we are entitled to evaluate its responses on the basis that they are \textit{bona fide} attempts at drawing valid conclusions, correctly answering questions, etc., as requested by user instructions.

\textbf{AV: Base LLMs can be steered to follow user instructions at inference time \cite{stolfo2025improvinginstructionfollowinglanguagemodels} and even generate long reasoning chains to derive correct answers \cite{venhoff2025basemodelsknowreason}.}

\textit{When these techniques are employed, one could argue that these models are in fact optimizing for the same normative qualities as instruct LLMs.}

Our position does not dispute the effectiveness of these techniques. In fact, much like instruction-tuning itself, we view these methods as potential ways to control for precisely the confounding variable we have identified.\footnote{In theory, few-shot demonstration \cite{10.5555/3495724.3495883} (including in-context examples of similar questions followed by correct answers) can also be adapted to specifically control for this confound. However, in practice, this method often fails as a control because it introduces its own confounding variables: e.g. the model may be learning only to mimic superficial patterns from the demonstrations, such as a specific output format or the statistical distribution of the labels \cite{min-etal-2022-rethinking}. Exactly how and why in-context demonstrations improves performance remains unclear and subject to ongoing study (\citealp{agarwal2024manyshot}; \citealp{nafar-etal-2025-learning}).} Our point remains that \textbf{whether the model is genuinely following instructions to reason and answer questions correctly is a confounding factor when assessing and making claims about its reasoning limitations, biases and processes}.

While this critique is therefore perfectly compatible with the existence of such techniques, the onus remains on the researcher to ensure that the instruction-following confound is sufficiently controlled such that they can draw valid conclusions about models' inherent reasoning abilities and limitations. As observed in Section \ref{sec:llm_reasoning_lit_review}, this is a key methodological consideration that has not been accounted for in existing studies and needs to be recognized in future work.

\textbf{AV: Benchmarking base LLMs is still a useful proxy for post-training reasoning abilities.}

\textit{The performance of base LLMs on reasoning benchmarks often correlates with its performance after post-training, making it a highly useful signal for tracking model development.}

We do not dispute the utility of assessing pre-trained base LLM performance on reasoning benchmarks as a convenient proxy for this purpose. However, our position does draw attention to the challenge in disentangling how much of the reasoning improvement observed after post-training is attributable to genuine improvements in reasoning capability versus improvements in instruction-following. This distinction is crucial in pinpointing what improves a model's inherent reasoning capabilities and diagnosing patterns of genuine reasoning errors in the training process.

\section{Conclusion and Recommendations}

In this paper, we have argued that evaluating the reasoning capabilities of base LLMs through prompting is methodologically problematic. Our position centers on a fundamental objective mismatch: unlike instruct LLMs, base LLMs are optimized only for linguistic plausibility, not for normative qualities such as correctness or validity. 

This mismatch has crucial implications. First, apparent ability of base LLMs to generate correct conclusions is purely incidental: as illustrated in our case study, these models indifferently generate both valid and invalid conclusions by simply conforming to a learned linguistic pattern. Second, this mismatch introduces an uncontrolled confound when evaluating base LLMs for reasoning capabilities: we cannot reliably distinguish \textit{bona fide} reasoning errors from outputs caused by a mere misalignment in objective (e.g., the model was not even following user instructions in attempting a \textit{correct} answer). These implications pose a serious challenge to the validity of claims about base LLMs' inherent reasoning limitations and biases, and the extent to which such claims can be generalized to instruct LLMs. 

On this basis, we recommend that future research prioritizes instruct LLMs in the evaluation of model reasoning, as their optimization for correctness and other desirable normative qualities provides a much stronger basis for identifying patterns of genuine reasoning errors or systematic biases. Existing findings on reasoning limitations of base LLMs should be critically re-examined and cannot be assumed to generalize to instruct LLMs. Furthermore, evaluation of a base LLM's reasoning must recognize and control for the inherent confound that the model may not be attempting to provide correct answers, valid conclusions, or otherwise successfully fulfill user instructions. 

\section*{Acknowledgements}
This work was supported by the UKRI AI Centre for Doctoral Training in Speech and Language Technologies (SLT) and their Applications funded by UK Research and Innovation [grant number EP/S023062/1]. For the purpose of open access, the author has applied a Creative Commons Attribution (CC BY) licence to any Author Accepted Manuscript version arising. We acknowledge IT Services at The University of Sheffield for the provision of services for High Performance Computing.

\bibliography{tacl2021}
\bibliographystyle{acl_natbib}








  

\end{document}